\documentclass{Interspeech2024}

\usepackage{hyperref}
\usepackage{todonotes}
\usepackage{xspace}
\usepackage{multirow}
\usepackage{ amssymb }
\usepackage{scrextend}
\usepackage{subfigure}
\usepackage{comment}

\newcommand{\newpar}{\vspace{1mm}\noindent\textbf}

\newcommand{\secref}[1]{\S\ref{#1}}
\usepackage{cite}
\graphicspath{{figures/}}

\def\ps@IEEEtitlepagestyle{%
  \def\@oddfoot{\mycopyrightnotice}%
  \def\@evenfoot{}%
}
\def\mycopyrightnotice{%
  {\footnotesize ACCEPTED AT INTERSPEECH 2024\hfill}% <--- Change here
  \gdef\mycopyrightnotice{}% just in case
}

\interspeechcameraready

\newcommand{\poli}{$^{\dagger}$}

\title{A Contrastive Learning Approach to Mitigate Bias in Speech Models}

\name[affiliation={\dagger}]{Alkis}{Koudounas}
\name[affiliation={\dagger}]{Flavio}{Giobergia}
\name[affiliation={\dagger}]{Eliana}{Pastor}
\name[affiliation={\dagger}]{Elena}{Baralis}

\address{
  \poli Politecnico di Torino, Turin, Italy
  }
\email{\{firstname.lastname\}@polito.it}


\begin{document}

\maketitle

\begin{abstract}
Speech models may be affected by performance imbalance in different population subgroups, raising concerns about fair treatment across these groups.
Prior attempts to mitigate unfairness either focus on user-defined subgroups, potentially overlooking other affected subgroups, or do not explicitly improve the internal representation at the subgroup level.
This paper proposes the first adoption of contrastive learning to mitigate speech model bias in underperforming subgroups.
We employ a three-level learning technique that guides the model in focusing on different scopes for the contrastive loss, i.e., task, subgroup, and the errors within subgroups.
The experiments on two spoken language understanding datasets and two languages demonstrate that our approach improves internal subgroup representations, thus reducing model bias and enhancing performance\footnote{This work has been accepted at Interspeech 2024.}.
\end{abstract}

\section{Introduction}
Ensuring balanced performance across diverse subgroups of data is a critical aspect of developing fair and unbiased speech models.
For instance, a model should perform similarly for all speakers, regardless of factors such as demographics and recording conditions. 
However, a growing body of work revealed speech models behave differently for different subpopulations
%~\cite{tatman2017effects, martin20_interspeech, 9610166, garnerin2021investigating, feng2021, zhang2022mitigating, liu2022, liu2022icassp, dheram2022, koudounas2023, zhang2023exploring, koudounas2024taslp}.
~\cite{tatman2017effects, martin20_interspeech, 9610166, feng2021, zhang2022mitigating, liu2022, liu2022icassp, dheram2022, koudounas2023, zhang2023exploring, koudounas2024taslp}.

This work proposes a novel approach to mitigate performance disparities in data subgroups by directly acting on the latent space representation of samples. To this end, we introduce CLUES, a Contrastive Learning framework for mitigating model biases towards UnderpErforming Subgroups.
Specifically, we leverage Contrastive Learning (CL), which has emerged as a significant advancement in representation learning~\cite{oord2018representation, wang2019multi}. 
The idea of CL is to learn representations that place similar samples close together and dissimilar ones further apart. 
We use this notion to guide the model in learning how to represent samples from the same subgroup close to each other.
The intuition is that refining the model representations at the subgroup level enables it to better capture their distinct characteristic, thus mitigating performance disparities.

CLUES employs a three-level contrastive learning loss, with
(i) a first contrastive term that operates at the task level, grouping together samples sharing the same class and separating different classes;  
(ii) a second term to map samples of the same subgroup close together in the embedding space while separating different subgroups; 
(iii) a third loss term that correctly predicted operates within each subgroup, aggregating samples while setting apart incorrect ones.
These losses guide the model in learning representations that capture different scopes, i.e., tasks, subgroups, and errors within subgroups, resulting in more informative embeddings.
While the first loss targets the overall model performance, the second and third losses aim to decrease performance disparities between subgroups, thus further enhancing model behavior.

Recent approaches for mitigation propose targeted data augmentation \cite{zhang2023exploring}, acquiring focused data \cite{koudounas2024icassp, dheram2022}, incorporating information from automatically identified subgroups at training time \cite{veliche2023improving}, or employing ad-hoc loss functions \cite{zhang2022mitigating}. 
However, these methods do not explicitly introduce mitigation by improving the latent representations.

Fairer representations via CL have been beneficial to address disparities for tasks such as text~\cite{chi2022conditional, shen2021contrastive} and image classification~\cite{hong2021unbiased},  machine translation~\cite{lee2023target}, and in pre-trained language models on downstream tasks~\cite{dong2023copt}.
However, these works address modalities different from speech. Moreover, most of these works focus on improving representations for known protected subgroups. 
However, disparities may occur for subpopulations that are unknown \textit{a priori}.
In response, we mitigate disparities for automatically identified subgroups. 

While multiple works adopt CL to train speech models \cite{al2021clar, ye2022cross, vaessen2024effect, han2021supervised}, the adoption of contrastive learning for fairness in the speech domain remains largely unexplored. 
To the best of our knowledge, our work is the first to integrate contrastive loss terms to obtain improved representations in speech models.

We experimentally evaluate our approach on two public datasets for intent classification,  \textsc{FSC}~\cite{fsc} in English and   \textsc{ITALIC}~\cite{koudounas23interspeech} in Italian, with the state-of-the-art transformer models wav2vec 2.0~\cite{Wav2Vec2} and XLS-R~\cite{xlsr_babu}.
The experimental results demonstrate that models trained with our contrastive learning schema develop richer representations that effectively reduce subgroup performance disparities and improve overall performance.
Specifically, we reduce the disparity in performance of the most underperforming subgroup by 66.9\% (15.5\%) for \textsc{FSC} (\textsc{ITALIC}) w.r.t. the baseline models.
We also observe an increase of 6.1\% (4.8\%) in the overall F1 Macro.

\section{Methodology}
We propose CLUES, a Contrastive Learning framework for mitigating biases toward UnderpErforming Subgroups. 
CLUES requires defining the subpopulations of interest to guide representation learning. 
Although CLUES is agnostic to the method adopted for defining subgroups, we discuss two possible approaches in \secref{sec:divexplorer}.
Based on the extracted subgroups, we present the three-step contrastive learning schema in \secref{sec:clues}.

\subsection{Subgroups identification}\label{sec:divexplorer}
In this work, we consider two possible subgroup extraction techniques: K-Means~\cite{Lloydkmeans} clustering and DivExplorer~\cite{divexplorerpaper}. 

\newpar{K-means clustering}. K-Means is a commonly adopted clustering algorithm, which assigns one of $K$ clusters to any sample. We consider each cluster as a subgroup. 
We apply the clustering algorithm to the latent representations of the input points, as extracted by the backbone model used. The cluster extraction is done at the beginning of each training epoch. 
In this way, the subgroups reflect the evolution of the latent representations throughout the training.

\newpar{DivExplorer}. DivExplorer~\cite{divexplorerpaper} leverages (typically interpretable) metadata to construct subgroups that satisfy a frequency threshold within the dataset. 
In the case of speech, these metadata may relate to speaker traits (e.g., gender, age, accent), recording conditions (e.g., utterance duration, noise levels, speaking rate), and task characteristics (e.g., the action, object, and location of the intent). 
DivExplorer introduces the concept of divergence as the difference in performance between a subgroup and the overall data. We use this notion to assign each point to the most divergent subgroup to which the sample belongs. 
In this way, we guide CLUES toward mainly improving representations of underperforming subgroups, i.e., the subgroups that diverge the most from the average model behavior. 

\begin{table*}[!ht]
\centering
\caption{
Mean $\pm$ std %and standard deviation 
of three runs on \textsc{FSC} and \textsc{ITALIC}. Comparison of \textit{original} fine-tuning, data augmentation~\cite{zhang2023exploring}, adversarial loss~\cite{zhang2022mitigating}, data acquisition~\cite{koudounas2024icassp}, and CLUES. For all metrics, higher is better. The best results are in \textbf{bold}, the second-best are \underline{underlined}. }
\label{tab:results}
\scalebox{0.98}{%
\begin{tabular}{lccccccc}
\toprule
\multicolumn{1}{c}{\textit{\textit{DS}}} 
    & \textit{\textit{Approach}} 
    & \textit{\textit{Subgroups}} 
    & \textit{\textit{Accuracy}} 
    & \textit{\textit{F1 Macro}} 
    & \textit{\textbf{\textit{$\Delta^-_{max}$}}} 
    & \textit{\textbf{\textit{$S$}}}
    & \textit{\textbf{\textit{$S^{\pm}$}}}
    % & \textit{\textbf{\textit{$\left\lVert E \right\rVert_{avg} \downarrow $}}}
    % & \textit{\textbf{\textit{$SE_{avg} \uparrow $}}}
    \\ \midrule
    
\multirow{8}{*}{\rotatebox[origin=c]{90}{\textsc{FSC}}} 
    & \texttt{w2v2-b} original 
    & -
    & 93.419\scriptsize$\pm$0.169 
    & 93.110\scriptsize$\pm$0.168 
    & -53.179\scriptsize$\pm$0.147 
    & 0.737\scriptsize$\pm$0.049
    & 0.318\scriptsize$\pm$0.084 \\

    & w/ data$++$~\cite{zhang2023exploring} 
    & -
    & 94.909\scriptsize$\pm$0.870 
    & 94.460\scriptsize$\pm$0.861 
    & -42.623\scriptsize$\pm$2.939
    & 0.757\scriptsize$\pm$0.031
    & 0.309\scriptsize$\pm$0.021 \\

    % & w/ data$_S^{++}$~\cite{ko2017study}
    % & \texttt{K-Means}
    % & 97.850\scriptsize$\pm$0.367 
    % & 97.589\scriptsize$\pm$0.650
    % & -37.569\scriptsize$\pm$8.680 
    % & 0.781\scriptsize$\pm$0.027
    % & 0.344\scriptsize$\pm$0.034 \\

    % & w/ data$_S^{++}$~\cite{ko2017study}
    % & \texttt{DivExplorer}
    % & 98.459\scriptsize$\pm$0.110
    % & 98.418\scriptsize$\pm$0.172
    % & \textbf{-15.512\scriptsize$\pm$0.561} 
    % & 0.754\scriptsize$\pm$0.035
    % & 0.331\scriptsize$\pm$0.039  \\
    
    & w/ adversarial~\cite{zhang2022mitigating}
    & \texttt{K-Means}
    & \underline{98.591\scriptsize$\pm$0.210}
    & 98.507\scriptsize$\pm$0.187
    & -26.141\scriptsize$\pm$0.117
    & 0.819\scriptsize$\pm$0.016
    & 0.401\scriptsize$\pm$0.017 \\

    & w/ adversarial~\cite{zhang2022mitigating}
    & \texttt{DivExplorer}
    & 98.486\scriptsize$\pm$0.105
    & 98.311\scriptsize$\pm$0.109
    & -24.512\scriptsize$\pm$0.145
    & 0.814\scriptsize$\pm$0.015
    & 0.389\scriptsize$\pm$0.021 \\

    & w/ data acquisition~\cite{koudounas2024icassp}
    & \texttt{K-Means}
    & 96.511\scriptsize$\pm$0.309 
    & 95.983\scriptsize$\pm$0.338 
    & -32.488\scriptsize$\pm$0.461
    & 0.749\scriptsize$\pm$0.048
    & 0.324\scriptsize$\pm$0.053 \\
    
    & w/ data acquisition~\cite{koudounas2024icassp}
    & \texttt{DivExplorer}
    & 96.719\scriptsize$\pm$0.215 
    & 96.054\scriptsize$\pm$0.274 
    & -22.692\scriptsize$\pm$0.316
    & 0.755\scriptsize$\pm$0.029
    & 0.334\scriptsize$\pm$0.032 \\
    
    & w/ CLUES 
    & \texttt{K-Means}
    & 98.567\scriptsize$\pm$0.145
    & \underline{98.514\scriptsize$\pm$0.141}
    & \underline{-21.410\scriptsize$\pm$0.393}
    & \underline{0.848\scriptsize$\pm$0.019}
    & \underline{0.516\scriptsize$\pm$0.015} \\

    & w/ CLUES 
    & \texttt{DivExplorer}
    & \textbf{98.789\scriptsize$\pm$0.104} 
    & \textbf{98.761\scriptsize$\pm$0.095} 
    & \textbf{-17.581\scriptsize$\pm$0.433}
    & \textbf{0.894\scriptsize$\pm$0.012}
    & \textbf{0.525\scriptsize$\pm$0.014} \\
    
    \midrule\midrule
    
\multirow{8}{*}{\rotatebox[origin=c]{90}{\textsc{ITALIC}}} 
    & \texttt{XLSR-300} original 
    & -
    & 75.711\scriptsize$\pm$0.360 
    & 73.218\scriptsize$\pm$0.329 
    & -47.541\scriptsize$\pm$0.789 
    & 0.319\scriptsize$\pm$0.064
    & -0.221\scriptsize$\pm$0.081 \\

    & w/ data$++$~\cite{zhang2023exploring}
    & -
    & 76.062\scriptsize$\pm$0.289
    & 73.361\scriptsize$\pm$0.771
    & -45.820\scriptsize$\pm$1.892
    & 0.323\scriptsize$\pm$0.099
    & -0.213\scriptsize$\pm$0.091 \\

    % & w/ data$_S^{++}$~\cite{zhang2023exploring}
    % & \texttt{K-Means}
    % & 78.012\scriptsize$\pm$0.450 
    % & 74.453\scriptsize$\pm$0.351 
    % & -44.810\scriptsize$\pm$2.349 
    % & 0.359\scriptsize$\pm$0.078
    % & -0.198\scriptsize$\pm$0.082 \\

    % & w/ data$_S^{++}$~\cite{zhang2023exploring}
    % & \texttt{DivExplorer}
    % & 78.071\scriptsize$\pm$0.532 
    % & 74.851\scriptsize$\pm$0.303 
    % & \underline{-40.498\scriptsize$\pm$1.710}
    % & 0.331\scriptsize$\pm$0.057
    % & -0.205\scriptsize$\pm$0.049 \\
    
    & w/ adversarial~\cite{zhang2022mitigating}
    & \texttt{K-Means}
    & 77.499\scriptsize$\pm$0.315
    & 75.014\scriptsize$\pm$0.437
    & -44.117\scriptsize$\pm$0.654
    & 0.447\scriptsize$\pm$0.098
    & -0.104\scriptsize$\pm$0.089 \\

    & w/ adversarial~\cite{zhang2022mitigating}
    & \texttt{DivExplorer}
    & 77.201\scriptsize$\pm$0.641
    & 74.840\scriptsize$\pm$0.527
    & -42.535\scriptsize$\pm$0.714
    & 0.461\scriptsize$\pm$0.086
    & -0.094\scriptsize$\pm$0.075 \\

    & w/ data acquisition~\cite{koudounas2024icassp}
    & \texttt{K-Means}
    & 76.308\scriptsize$\pm$0.512
    & 74.016\scriptsize$\pm$0.505
    & -41.918\scriptsize$\pm$0.672
    & 0.375\scriptsize$\pm$0.024
    & -0.210\scriptsize$\pm$0.027 \\
    
    & w/ data acquisition~\cite{koudounas2024icassp}
    & \texttt{DivExplorer}
    & 77.510\scriptsize$\pm$0.441
    & 75.201\scriptsize$\pm$0.384
    & \underline{-41.005\scriptsize$\pm$0.510}
    & 0.389\scriptsize$\pm$0.019
    & -0.197\scriptsize$\pm$0.022 \\
    
    & w/ CLUES 
    & \texttt{K-Means}
    & \textbf{80.561\scriptsize$\pm$0.554} 
    & \underline{76.104\scriptsize$\pm$0.317}
    & -43.010\scriptsize$\pm$0.892
    & \underline{0.512\scriptsize$\pm$0.034}
    & \underline{0.214\scriptsize$\pm$0.028} \\

    & w/ CLUES 
    & \texttt{DivExplorer}
    & \underline{79.230\scriptsize$\pm$0.810}
    & \textbf{76.721\scriptsize$\pm$0.201} 
    & \textbf{-40.150\scriptsize$\pm$0.963}
    & \textbf{0.539\scriptsize$\pm$0.025}
    & \textbf{0.241\scriptsize$\pm$0.023} \\
    
    \bottomrule
\end{tabular}
}
\end{table*}
\begin{table}[!ht]
\centering
\caption{
Ablation study on $\mathcal{L}_t$, $\mathcal{L}_s$, and $\mathcal{L}_e$. The best results are in \textbf{bold}. Subgroups extracted with DivExplorer.}
\label{tab:ablation}
\scalebox{0.93}{%
\begin{tabular}{lcccccc}
\toprule
\multicolumn{1}{c}{\textit{\textit{DS}}} 
    & \textit{\textit{Approach}} 
    %& \textit{\textit{Accuracy $\uparrow$}} 
    & \textit{\textit{F1 Macro}} 
    & \textit{\textbf{\textit{$\Delta^-_{max}$}}} 
    & \textit{\textbf{\textit{$S $}}}
    & \textit{\textbf{\textit{$S^{\pm}$}}}
    % & \textit{\textbf{\textit{$\left\lVert E \right\rVert_{avg} \downarrow $}}}
    % & \textit{\textbf{\textit{$SE_{avg} \uparrow $}}}
    \\ \midrule
    
\multirow{7}{*}{\rotatebox[origin=c]{90}{\textsc{FSC}}} 

    & \texttt{w2v2-b}
    %& 93.419
    & 93.110
    & -53.179
    & 0.737
    & 0.318 \\
    
    & w/ $\mathcal{L}_t$ 
    %& 98.214
    & 98.105
    & -48.714
    & 0.759
    & 0.309 \\

    & w/ $\mathcal{L}_s$ 
    %& 98.441
    & 98.434
    & -26.551
    & 0.848
    & 0.419 \\

    & w/ $\mathcal{L}_t + \mathcal{L}_s$ 
    %& 98.438
    & 98.430
    & -33.124
    & 0.814
    & 0.371 \\

    & w/ $\mathcal{L}_s + \mathcal{L}_e$ 
    %& 98.455
    & 98.452
    & -19.112
    & 0.861
    & 0.501 \\

    & w/ $\mathcal{L}_s + \mathcal{L}_e^*$ 
    %& 98.119
    & 98.106
    & -20.014
    & 0.865
    & 0.487 \\
    
    & w/ CLUES 
    %& \textbf{98.789} 
    & \textbf{98.761} 
    & \textbf{-17.581}
    & \textbf{0.894}
    & \textbf{0.525} \\
    
    \midrule\midrule
    
\multirow{7}{*}{\rotatebox[origin=c]{90}{\textsc{ITALIC}}} 

    & \texttt{XLSR-300}  
    %& 75.711
    & 73.218
    & -47.541
    & 0.319
    & -0.221 \\
    
    & w/ $\mathcal{L}_t$ 
    %& 78.971
    & 76.075
    & -49.540
    & 0.346
    & -0.222 \\

    & w/ $\mathcal{L}_s$ 
    %& 78.995
    & 76.326
    & -45.391
    & 0.452
    & -0.196 \\

    & w/ $\mathcal{L}_t + \mathcal{L}_s$ 
    %& 78.987
    & 76.279
    & -47.443
    & 0.439
    & -0.208 \\

    & w/ $\mathcal{L}_s + \mathcal{L}_e$ 
    %& 79.022 
    & 76.620 
    & -43.288
    & 0.490
    & 0.215 \\

    & w/ $\mathcal{L}_s + \mathcal{L}_e^*$ 
    %& 78.945
    & 76.319
    & -43.329
    & 0.499 
    & 0.123 \\
    
    & w/ CLUES 
    %& \textbf{79.230} 
    & \textbf{76.721} 
    & \textbf{-40.150}
    & \textbf{0.539}
    & \textbf{0.241} \\
    
    \bottomrule
\end{tabular}
}
\end{table}

\subsection{CLUES}\label{sec:clues}
\begin{figure}
    \centering
    \includegraphics[width=0.9\linewidth]{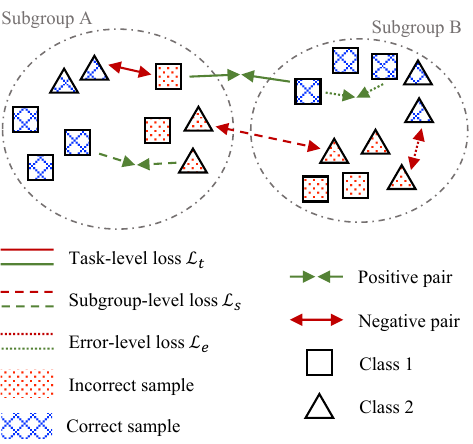}
    \caption{Summary of the action of the three contrastive loss terms on a toy example, comprised of 2 subgroups (A, B) and a binary classification task (square/triangle). }
    \label{fig:losses}
\end{figure}

Our approach aims to \textit{hint} the model toward learning embeddings that simultaneously achieve: (i) the separation of samples according to a classification task (e.g., intent classification), (ii) the separation of samples according to the identified subgroups, and (iii) an equitable representation of over- and under-performing subgroups within the classification task. 
To this end, CLUES consists of three separate contrastive learning levels: task, subgroup, and error. At each level, we employ the multi-similarity (MS) loss~\cite{wang2019multi} to selectively contrast sample pairs based on their affinity. 
This loss has already proven to be effective in speech tasks~\cite{laquatra2024}.
For any sample (referred to as anchor in CL), we identify positive and negative samples based on the various levels' criteria. The contrastive loss works toward getting positive samples closer to the anchor and negative samples further away from it.
For each level, the corresponding loss is computed as the sum of positive and negative terms:
\begin{equation}
    \begin{aligned}
        \mathcal{L}_{MS} = \dfrac{1}{m} \sum_{i=1}^m  
        \dfrac{1}{\alpha} \log [ 1 + \sum_{p \in \mathcal{P}_i} e^{-\alpha (S_{ip} - \lambda)} ] \\
        + \dfrac{1}{\beta} \log [ 1 + \sum_{n \in \mathcal{N}_i} e^{\beta (S_{in} - \lambda)} ] 
    \end{aligned}
\end{equation}
Where $m$ is the batch size, $\mathcal{P}_i$ and $\mathcal{N}_i$ denote the sets of positive and negative samples for the anchor $i$, $S_{ip}$ and $S_{in}$ are the similarities between $i$ and its positive/negative pairs, and $\alpha, \beta, \lambda$ control pair weighting. The three loss terms introduced, which are summarized in Figure~\ref{fig:losses}, are described as follows.

\newpar{Task-level Contrastive Learning.} We introduce a first MS contrastive loss function focused on the classification task level $\mathcal{L}_t$. This loss groups samples sharing the same class and separates samples belonging to different classes, and aims at improving the separability of the samples in the downstream task.

\newpar{Subgroup-level Contrastive Learning.} 
We use a second MS contrastive loss function $\mathcal{L}_{s}$ to guide the learning of subgroup-level representations. We leverage the subgroups identified within the data, as described in Section~\secref{sec:divexplorer}. 
In particular, we choose as positive pairs points belonging to the same subgroup, and as negative pairs points belonging to different ones. 
This encourages an internal representation that is aware of the identified subpopulations, not only of the task being addressed. 

\newpar{Error-level Contrastive Learning.} Finally, we introduce a contrastive loss term $\mathcal{L}_{e}$ that considers intra-subgroup errors on the classification task.
The outcome of a model for each sample can be represented as a binary (correct/incorrect) result.
Within each subgroup, we define as positive the pairs of samples that have obtained the same outcome, and as negative the ones with different outcomes. 
This introduces a bipartition within each subgroup: one of the partitions contains all correctly predicted samples, and the other contains all incorrectly predicted ones.
As samples get predicted correctly, the $\mathcal{L}_e$ loss moves them toward the correctly predicted partition.

\newpar{Final Loss.} We define our overall training objective as the aggregation of these multi-similarity losses, along with a conventional classification loss $\mathcal{L}_{cls}$ (e.g., cross-entropy): 

\begin{equation}
    \mathcal{L} = \mathcal{L}_{cls} + \lambda_t \mathcal{L}_{t} + \lambda_s \mathcal{L}_{s} + \lambda_e \mathcal{L}_{e}
\end{equation}

\noindent
where $\lambda_t$, $\lambda_s$ and $\lambda_e$ are coefficients that regulate the relative importance of each loss term. We identify the best values for these coefficients with a tuning phase using a validation set.

\begin{figure*}[t]
  \centering
  \includegraphics[width=0.9\linewidth]{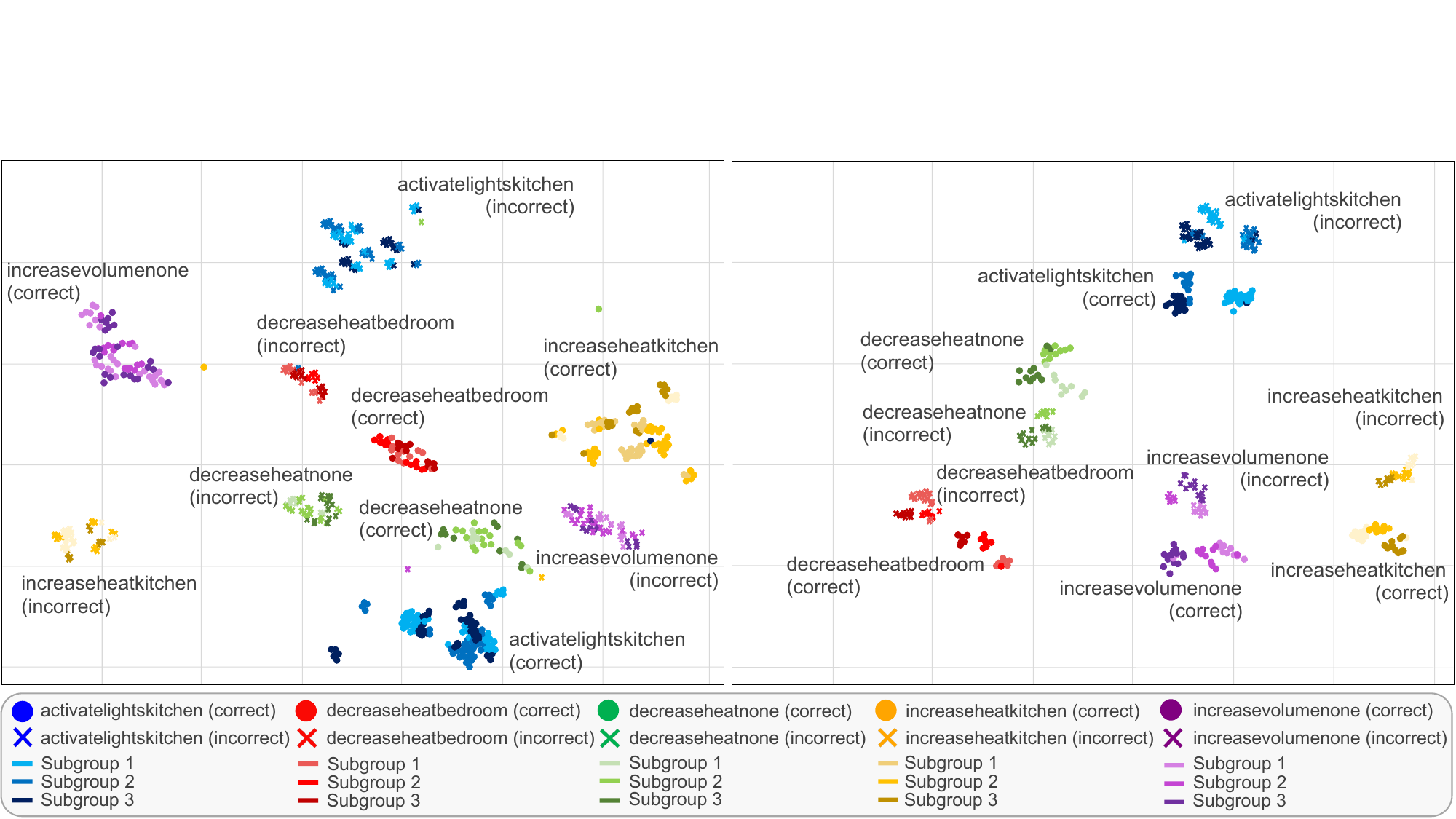}
  \caption{\textsc{FSC}. 5 most-frequent intents (main color) and 3 most frequent subgroups (shades of the same color). t-SNE visualization 
  of the original model (left) and CLUES (right). Correct samples reported as circles, incorrect ones as crosses. Best viewed in color.
  }
  \label{fig:fsc-tsne}
\end{figure*}

\section{Experimental Setup}
In this section, we detail the setup used for the experiments\footnote{\texttt{https://github.com/koudounasalkis/CLUES}}.

\newpar{Datasets.} We evaluated our approach on two public intent classification datasets, \textsc{FSC}~\cite{fsc} for English and \textsc{ITALIC}~\cite{koudounas23interspeech} for Italian. 
\textsc{FSC} includes 30,043 utterances annotated with action, object, and location defining intents, while \textsc{ITALIC} contains 16,521 samples with action and scenario denoting intents. Both datasets split speakers across train, validation, and test sets.

\newpar{Metadata.}  We considered demographic, speaking and recording conditions, and intent-related metadata, following the metadata-enrichment proposed in~\cite{koudounas2023}. 
When using DivExplorer, we explored all subgroups with a minimum frequency of $0.03$, while for K-Means we considered $K$=10 for \textsc{ITALIC} and $K$=20 for \textsc{FSC}. These configurations have been found to achieve the best performance on the target datasets~\cite{koudounas2024icassp}.

\newpar{Models.} We fine-tuned the pre-trained wav2vec 2.0~\cite{Wav2Vec2} base and multilingual XLS-R~\cite{xlsr_babu} models on the \textsc{FSC} and \textsc{ITALIC} datasets, respectively. The pre-trained checkpoints were obtained from the Hugging Face hub~\cite{huggingface} and served as our baselines.
We followed fine-tuning procedures from relevant literature~\cite{superb}. 
We adhere to standard procedures in CL, selecting positive and negative sample pairs within each batch to optimize model performance.
Further details about models, hyperparameters, and fine-tuning are available in the project repository.

\newpar{Metrics.} We assessed the overall model performance with accuracy and macro F1 score.
We also considered the highest negative subgroup divergence ($\Delta^-_{max}$), where the divergence $\Delta$ of a subgroup is the difference between that subgroup accuracy and the overall one. 
$\Delta^-_{max}$ evaluates how well the model can reduce differences in performance between subgroups and thus mitigate bias. 
Finally, we evaluated the quality of the latent space representation in terms of Silhouette~\cite{rousseeuw1987silhouettes}, which quantifies intra-subgroup cohesion and inter-subgroup separation.
We measured both the Silhouette w.r.t. the adopted subgroups ($S$), and the Silhouette w.r.t. the partitions of correctly/incorrectly predicted samples within the subgroups ($S^\pm$). 

\newpar{Baselines.} 
We evaluated CLUES against several competitors. We explored a standard data augmentation scenario~\cite{zhang2023exploring}.
Additionally, following~\cite{zhang2022mitigating}, we experimented with an extra adversarial loss that aims to predict whether an utterance belongs to an underperforming subgroup or not\footnote{While the authors of~\cite{zhang2022mitigating} employ an extra loss for distinguishing between native and non-native speakers, we propose to discern utterances belonging to underperforming and non-underperforming subgroups.}.
Finally, we included a baseline inspired by the work of~\cite{koudounas2024icassp}, simulating a scenario where additional data from the same distribution as the existing dataset is available. In this baseline, samples from underperforming subgroups are selectively added. 
For \cite{zhang2022mitigating}, \cite{koudounas2024icassp}, and CLUES, the subgroups are identified using both K-means and DivExplorer. 

\section{Experimental Results}
We report the main experimental results in Table \ref{tab:results}.
The outcomes obtained are consistent across the two datasets.
Both K-Means and DivExplorer enabled CLUES to perform well in general, demonstrating its ability to deal with different subgroup extraction techniques.
Since DivExplorer-based results are generally better, we focus the discussion on this approach.
The results show that CLUES achieved the best overall model performance in terms of both accuracy and F1 score.
On top of improving the overall performance, we also observed a decrease in maximum negative divergence. 
For instance, on \textsc{FSC}, the most underperforming subgroup deviates by 53.2 accuracy points from the overall performance. CLUES brings the divergence of the worst-performing subgroup down to -17.6\% instead. 
This decrease also surpasses the one obtained with the acquisition of new, targeted data. 
We mainly attribute this improvement to $\mathcal{L}_s$ and $\mathcal{L}_e$, which focus on reducing the intra-subgroup dispersion. 
We used the Silhouette to measure the quality of the latent space representations. 
Also in these terms, CLUES achieved by far the best performance for both $S$ and $S^\pm$, thanks to the $\mathcal{L}_s$ (for $S$) and $\mathcal{L}_s + \mathcal{L}_e$ terms (for $S^\pm$).

\newpar{Ablation Study.}
We conducted an ablation study to quantitatively assess the impact of each proposed loss term.
%of the proposed loss terms.
Table \ref{tab:ablation} summarizes these results.
All loss terms produce an improvement in accuracy and F1 score over the original model. 
The best improvement occurs when all loss terms are combined. 
The reduction in divergence ($\Delta^-_{max}$) mainly occurs when the $\mathcal{L}_s$ term is introduced. 
Indeed, this term acts by grouping together points belonging to the same subgroup, thus producing more easily separable latent representations. 
We observed an additional decrease in divergence and an improvement in terms of overall performance when adding $\mathcal{L}_e$.
For completeness, we additionally explored the option where the error-level loss merges correct and incorrect points together (instead of separating them). We refer to this alternative loss as $\mathcal{L}^{*}_e$.
The results with $\mathcal{L}^{*}_e$ are still satisfactory, indicating that the intra-subgroup action is still effective.
However, $\mathcal{L}_{e}$ has a better effect on performance w.r.t. $\mathcal{L}^{*}_{e}$.
Finally, as expected, we note how the introduction of the error-level produced the largest improvement in terms of $S^{\pm}$.

\newpar{Qualitative Analysis.}
We provide a qualitative analysis of the effect of CLUES on latent representations.
Figure \ref{fig:fsc-tsne} presents a t-SNE~\cite{van2008visualizing} visualization that compares the original (wav2vec 2.0) vector space against the CLUES-tuned approach.
We use the \textsc{FSC} dataset and select the 5 most frequent intents. 
For each intent, we visualize the samples belonging to the three most frequent subgroups.
As can be expected, both approaches tend to cluster together points sharing the same target intent.
However, the original embedding produces generally less cohesive clusters, since correctly and incorrectly predicted samples belonging to the same intent are not placed close together. 
Additionally, points belonging to the same subgroup are not necessarily close together in the original space. 
Instead, the contrastive terms introduced by CLUES produce more consistent groups of samples according to both the intent class and the subgroups.
We argue that this increase in group cohesion is the reason behind the observed general improvement in performance.
For instance, the samples for the \textit{increase heat kitchen} intent (in yellow) are particularly spread in the original representation -- with correct and incorrect samples being far away from one another and with little separation between subgroups.
By contrast, the CLUES-tuned version groups all relevant samples together into a more cohesive cluster on 3 levels. On a broader level, all points belonging to the intent are close together (e.g., all yellow points).
Within each intent, we can identify two partitions for correctly and incorrectly predicted samples (top and bottom of the yellow cluster).
Finally, within each partition, the three subgroups form three cohesive sub-clusters.

% \newpar{Discussion.} The empirical results showed that CLUES obtains a general improvement in model performance and a reduction in subgroup divergence, outperforming competitors.
% The ablation study confirmed that the loss terms act as desired on the various evaluation metrics.
% As expected, the combination of all loss terms produced the best results across all metrics.
% Finally, we verified that the latent representations obtained with CLUES are more consistent with the desired behavior, as shown with quantitative (via Silhouette) and qualitative (via t-SNE) results. 

% \section{Conclusions}
% We introduced CLUES, a contrastive learning framework to mitigate biases in underperforming subgroups. 
% Through a multi-level contrastive schema, CLUES steers the model towards a balanced understanding of both the classification task and underperforming subgroup modeling. 
% Experimental results demonstrate that our approach reduces subgroup disparities while improving overall model performance.
 
% \newpar{Limitations.} In this work, we presented experimental results only on intent classification. However, our methodology could be readily extended to other classification tasks. 
% We additionally plan on extending CLUES to other supervised tasks, such as Automatic Speech Recognition.

\section{Discussion}
We introduced CLUES, a contrastive learning framework to mitigate biases in underperforming subgroups. 
Through a multi-level contrastive schema, CLUES steers the model towards a balanced understanding of both the classification task and underperforming subgroup modeling. 
The results showed that CLUES obtains an improvement in model performance and a reduction in subgroup divergence, outperforming competitors.
The ablation study confirmed that the loss terms act as desired on the evaluation metrics.
As expected, the combination of all loss terms produced the best results across all metrics.
Finally, we verified that the latent representations obtained with CLUES are more consistent with the desired behavior, as shown with quantitative (via Silhouette) and qualitative (via t-SNE) results. 
 
\newpar{Limitations.} In this work, we presented experimental results only on intent classification. However, our methodology could be readily extended to other classification tasks. 
We additionally plan on extending CLUES to other supervised tasks, such as Automatic Speech Recognition.

\section{Acknowledgments}
The authors thank Giuseppe Averta and Moreno La Quatra for the useful discussions, and Sara Papi for her valuable support during the paper writing.
This work is partially supported by the FAIR - Future Artificial Intelligence Research (PIANO NAZIONALE DI RIPRESA E RESILIENZA (PNRR) – MISSIONE 4 COMPONENTE 2, INVESTIMENTO 1.3 – D.D. 1555 11/10/2022, PE00000013) and the spoke ``FutureHPC \& BigData'' of the ICSC - Centro Nazionale di Ricerca in High-Performance Computing, Big Data and Quantum Computing, both funded by the European Union - NextGenerationEU. 
This manuscript reflects only the authors' views and opinions, neither the European Union nor the European Commission can be considered responsible for them. 

\bibliographystyle{IEEEtran}
\bibliography{camera_ready}

\end{document}